\documentclass{article}
\usepackage{ijcai20}

\usepackage{times}

\usepackage{soul}
\usepackage{url}
\usepackage[utf8]{inputenc}
\usepackage[small]{caption}
\usepackage{graphicx}
\usepackage{amsmath}
\usepackage{multirow} 
\usepackage{booktabs}
\usepackage{xcolor}
\usepackage{subfig}
\usepackage{floatrow} 

\title{Classifying Wikipedia in a fine-grained hierarchy~: what graphs can contribute}

 \author{
 Tiphaine Viard$^{1,2}$\footnote{Contact Author}\and
 Thomas McLachlan$^2$\and
 Hamidreza Ghader$^{2,3}$\And
 Satoshi Sekine$^2$\\
 \affiliations
 $^1$Laboratoire d'Informatique de Paris Nord (LIPN),  Villetaneuse, France\\
 $^2$Language Information Access Team, Riken AIP, Tokyo, Japan\\
 $^3$ILPS, Amsterdam University
 \emails
 \{thomas.mclachlan, hamidreza.ghader,  satoshi.sekine\}@riken.jp,
 viard@lipn.univ-paris13.fr
}

\begin{document}

\maketitle

\begin{abstract}

Wikipedia is a huge opportunity for machine learning, being the largest semi-structured base of knowledge available. Because of this, many works examine its contents, and focus on structuring it in order to make it usable in learning tasks, for example by classifying it into an ontology.

Beyond its textual contents, Wikipedia also displays a typical graph structure, where pages are linked together through citations.
In this paper, we address the task of integrating graph ({\em i.e.} structure) information to classify Wikipedia into a fine-grained named entity ontology (NE), the {\em Extended Named Entity} hierarchy.

To address this task, we first start by assessing the relevance of the graph structure for NE classification.
We then explore two directions, one related to feature vectors using graph descriptors commonly used in large-scale network analysis, and one extending flat classification to a weighted model taking into account semantic similarity.
We conduct at-scale practical experiments, on a manually labeled subset of 22,000 pages extracted from the Japanese Wikipedia.

Our results show that integrating graph information succeeds at reducing sparsity of the input feature space, and yields classification results that are comparable or better than previous works. 
\end{abstract}
\section{Introduction}

{\em Named Entities} (NEs) concisely represent  the knowledge of people, things, and events, among others.
Having a hierarchy of such Named Entities, and being able to classify textual data into it, is crucial in order to structure knowledge, and as such make it reusable as a data source for applications. 

Classifying Wikipedia into a set of defined classes has been addressed in several works, see for instance~\cite{torisawa2007exploiting,dakka2008augmenting,nothman2013learning}.
However, these studies typically rely on a coarse hierarchy of named entities, not exceeding a few tens of types.
Coarse-grained classification is commonly used because it allows the researchers to have enough data points for each class, thus countering the data sparseness issue arising when applying supervised machine learning methods.
For example, articles such as ``Japan", ``Tokyo", and ``Tokyo Museum of Modern Art" may be classified in different fine-grained types, but all fall into a larger \texttt{Location} type.
Another challenge is that that fine-grained types are often overlapping.
For example, ``Tokyo Museum of Modern Art" can be classified as \texttt{Location} as well as \texttt{Culture}.
However, using a fine-grained hierarchy yields the benefit of precision, that is of crucial importance in many machine learning applications. The large amounts of data present in Wikipedia allows such precision, and as such fine-grained entity hierarchies are gaining more and more attention. In this paper, we rely on the {\em Extended Named Entity} hierarchy (ENE) as defined by~\cite{sekine2008extended} to classify articles of the online encyclopaedia Wikipedia.

Moreover, in many contexts, including Wikipedia, there is inherent structure to the data: thematically similar pages are likely to refer to each other, data is organized into categories~\cite{wu2008automatically}, etc.
This structure can be naturally modelled in the form of a graph, {\em i.e.} a set of nodes and a set of links between those nodes.

The Wikipedia graph has been studied in many contexts, and it has been shown to exhibit traits of a real-world graph structure~\cite{tsourakakis2009doulion}.
Typical applications of the graph analysis of Wikipedia include trust assessment~\cite{lucassen2010trust}, key page identification~\cite{hinnosaar2019wikipedia}, authority figure identification~\cite{kittur2007he}, or knowledge base construction~\cite{mahdisoltani2013yago3,lehmann2015dbpedia}.
Exploiting the graph structure can help classification, by identifying Wikipedia pages that describe the same roles, even though their contents are widely different: this might be the case, for example, of two head of states in different countries.

We argue in this paper that the underlying graphs of Wikipedia can aid classification, and in particular fight class imbalance; the gist behind this idea is that even for an ENE class with few annotated examples, we can rely on the rich underlying structure of the graph.
To address this, in this paper, we propose a hybrid classifier that takes into account the structure of Wikipedia as well as the natural language features devised in~\cite{suzuki2016fine}.

We are interested in predicting the named entity corresponding to a Wikipedia page; in other words, we are interesting in performing a classification task, mapping the Wikipedia pages into one of the categories of the ENE hierarchy. 
More specifically, in order to capture the structure of exchanges, we use the GraphSAGE learning algorithm~\cite{hamilton2017inductive}. Graph neural network models have known a recent development, and have shown to outperform the state-of-the-art methods in many contexts~\cite{kipf2016semi,hamilton2017inductive,schlichtkrull2017modeling}.

After assessing the relevance of the graph structure for the classification of Wikipedia in a fine-grained hierarchy, we explore different ways by which the graph structure can be integrated. One of them is the integration of graph-theoretical features, based on notions commonly used in real-world complex network analysis, whereas the other focuses on extending graph neural networks models, in particular by introducing a new aggregator for GraphSAGE, \textsc{Wmean}, that relies on a weighted mean of the vector representations of neighbours of a page.

Our work shows, in particular, that:
\begin{itemize}
    \item The underlying graph structure of Wikipedia, is helpful for natural language processing classification tasks, in particular in fighting class imbalance;
    \item That the ENE structure is hierarchical in nature, and that encoding this structure in the model is beneficial; 
    \item That in a context where deep learning models typically rely on large swaths of annotated data, graph features and models offer an alternative requiring less data.
\end{itemize}

In Section 2, we discuss the related works, before exploring in Section 3 the relevance of graph information for classification and defining the two directions mentioned above. Section 4 details our problem setting, and we present at-scale experiments in Section 5. We end our paper with perspectives for future work, in Section 6.

\section{Related Work}



In many contexts, Wikipedia can be considered as a knowledge base.
This has many advantages: the huge amount of data allows for cross-referencing information, data is typically available in multiple languages, and the data is semi-structured, since every page is categorised. 
However, as~\cite{zesch2007analysis} shows, the Wikipedia Category Graph alone is typically user-contributed, and as such is not necessarily consistent from one page to another, hindering any classification task.

{\bf Named Entity Classification.} On the other hand, expert-crafted ontologies such as CyC are curated and provide a consistent basis for classification.
However, to keep up-to-date with new concepts and retain their genericity, they typically comprise of few classes~\cite{fensel2001ontologies}, or are quickly outdated.

Fine-grained ontologies such as the {\em Extended Named Entities} ontology introduced by Sekine~\shortcite{sekine2008extended} come as a solution to both problems: being fine-grained, they capture the subtleties in the data, and being in part built from real-world datasets (such as a corpus of newspaper articles), they are up-to-date with current knowledge.
Yogatama et al.~\shortcite{yogatama2015embedding} give another example of such fine-grained ontology.
In this paper, we focus on the ENE ontology introduced by Sekine~\shortcite{sekine2008extended}, as it is adapted to the Japanese language.

The work of~\cite{suzuki2016fine} uses a Skip-gram model on the hypertext links to learn feature vectors, and as such is a first attempt at taking into account the structure induced by hyperlinks. In their paper,~\cite{shavarani2019multi} provide F1-score results for many baselines, on a different dataset extracted from the japanese Wikipedia, using the same ENE structure. It is, after the work of~\cite{suzuki2016fine}, the closest work from this paper.

{\bf Graph neural networks.} Learning on graph-based data has been a recent development in machine learning, in order to take into account the rich structure of real-world datasets.
Since first work defining a {\em graph neural network model}~\cite{scarselli2008graph}, a large number of models have emerged.

Some of the recent advances in graph neural networks have been loosely inspired by existing natural language processing methods, as for example with \textsc{Node2Vec}~\cite{grover2016node2vec}, based on \textsc{Word2Vec}; in turn, ~\cite{kipf2016semi} introduced GCN, which forms the basis for many current graph neural networks, such as \textsc{GraphSAGE}~\cite{hamilton2017inductive}.
These approaches heavily rely on random walks on the graph (or on subgraphs) to gain insight on nodes that are both close and far in the graph, without being intractable in terms of computation~\cite{perozzi2014deepwalk}. 

{\bf Wikipedia as graph.} Conversely, the many graphs structures in Wikipedia have been thoroughly studied~\cite{suh2007us,wu2011characterizing}, in applications related to natural language processing~\cite{aouicha2016derivation,zesch2007analysis}, or in large-scale graph analysis~\cite{brandes2009network}.
In all cases, the most common graph studied is the document-to-document graph~\cite{brandes2009network}, as well as the {\em Wikipedia Category Graph}~\cite{wu2011characterizing}.
Finally, some authors also consider the author-document graph, however since this graph is bipartite there is a lack of metrics to analyse it, and as such it has collected less attention.
In this paper, we will only consider the document-to-document graph.

\section{Graphs for ENE classification}
\subsection{Data}
    \label{sec:data}
    First of all, let us briefly present the dataset that we use in this paper.
    We focus on a dataset extracted from the online encyclopedia Wikipedia in the Japanese language.
    
    In the following, we model Wikipedia as a graph $G=(V,E)$, where $V$ is the set of nodes, and $E\subseteq V\otimes V$ is the set of edges without order or duplicates, {\em i.e.} for all $(u,v)\in E$, $u\ne v$ and $(u,v)=(v,u)$.
Elements of $V$ are pages from Wikipedia, and a link $(u,v)$ between page $u$ and $v$ means there is a citation between pages $u$ and $v$.
    
    This subset encompasses $22,639$ pages that were manually annotated into the $195$ classes of the ENE structure.
    Among those nodes, there are about $2.5\cdot 10^{6}$ links, each of those links representing one (or more) hyperlinks between the two linked pages.
    We use this dataset primarily as a means of comparison to recent works, and in particular~\cite{suzuki2016fine}.

   \subsection{Graph modelling for ENE prediction}
\label{sec:exps}
Let us explain the expected advantage of graph modelling in order to classify Wikipedia pages into the ENE hierarchy.

\begin{figure*}
    \centering
    \includegraphics[width=0.99\linewidth]{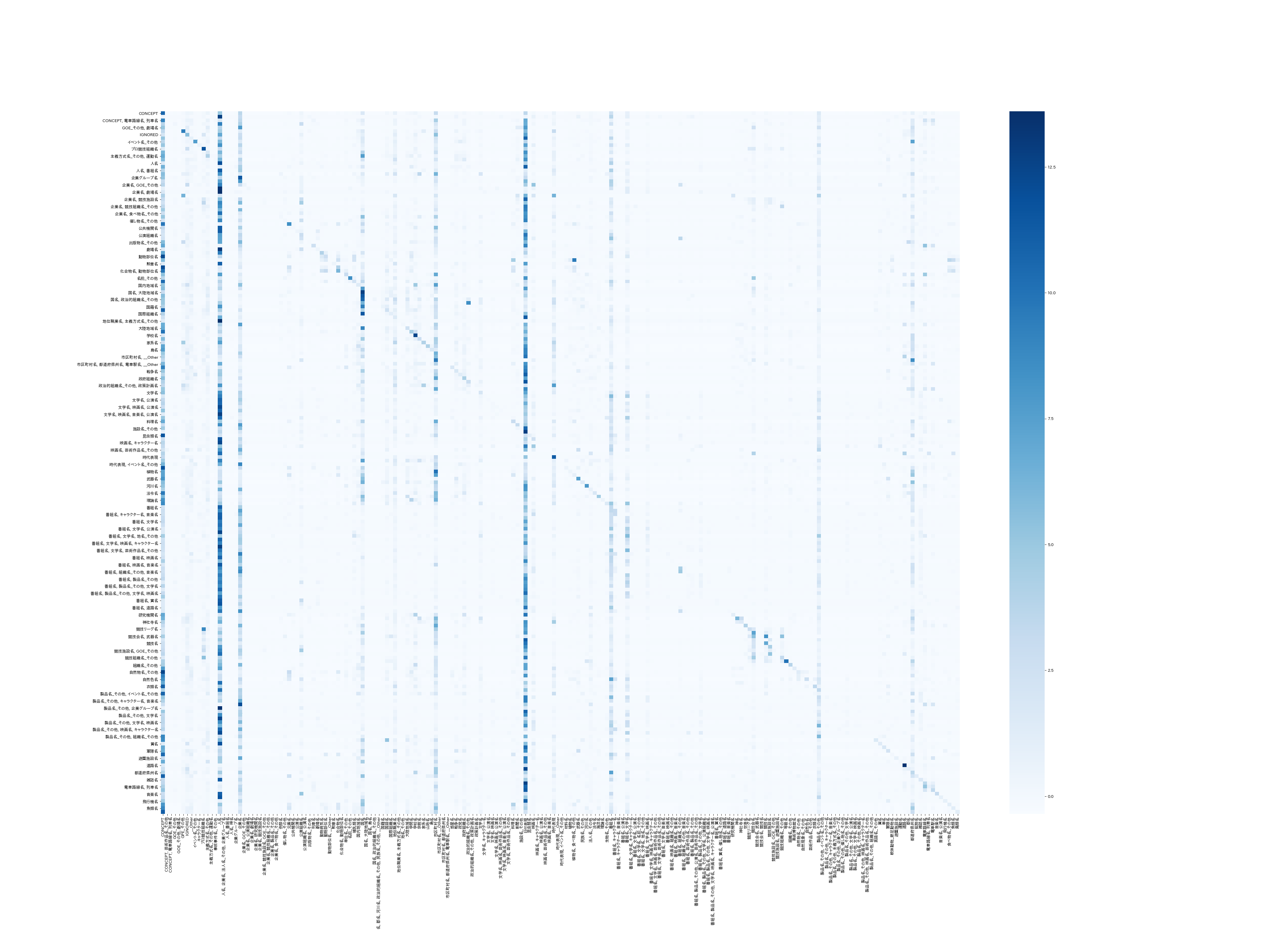}\\
    \caption{Heatmap representing the proportion of ENE $j$ in the neighbourhood of nodes having ENE $i$. ENEs in each row and column are sorted accroding to their order in the ENE hierarchy, which means that closely related ENEs are close to each other. We normalize such that for all row $i$, $\sum_{j} A_{ij}$ is constant. Notice that this makes the matrix non-symmetric.}
    \label{fig:ene_neighbourhood}
\end{figure*}

Figure~\ref{fig:ene_neighbourhood} displays a heat map with each row $i$ (column $j$) being ENEs, and where each element $A_{i,j}$ of the matrix is coloured according to the proportion of ENE $j$ in the neighbourhoods of nodes having ENE $i$.

\begin{figure*}
        \centering
        \subfloat[Distribution of the number of nodes (pages) per ENE. The most prevalent one, Person, encompasses around 4000 nodes. \label{fig:dist-enes-a}]{\includegraphics[width=0.48\linewidth]{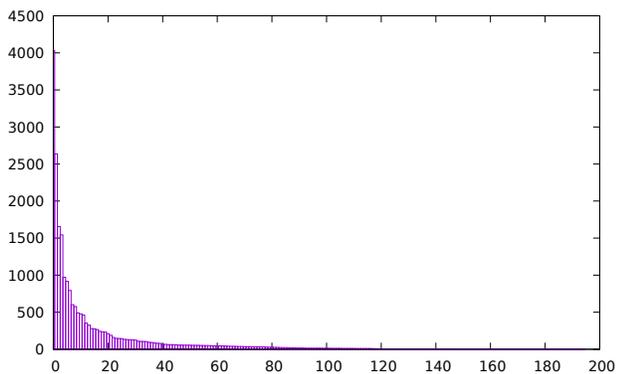}}\hfill
        \subfloat[Distribution of the number of ENEs in the neighbourhood of each node. \label{fig:dist-enes-b}]{\includegraphics[width=0.48\linewidth]{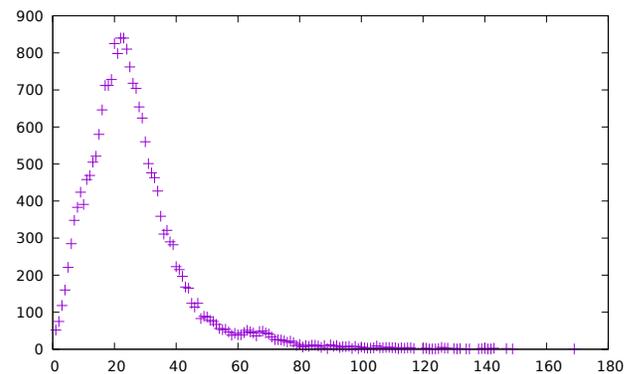}}\caption{Distributions related to the number of ENEs, demonstrating how taking into account the graph information can fight class imbalance.}\label{fig:dist-enes}
\end{figure*}

From this figure, we can derive different informations; first of all, there are "star" ENEs, that are highly represented in any node's neighbourhood.
Those typically correspond to Person, Country, Date, {\em i.e.} ENEs that are so prevalent that they are likely to be present in any page.
However, besides these "star" ENEs, we notice that (i) the diagonal is highlighted, meaning that if a node $u$ has ENE $i$, there likely exists (at least) a node $v$ linked to $u$ with the same ENE $i$, and (ii) we observe the formation of clusters (rectangles in the matrix).
These clusters, given the order imposed on the rows and columns of the matrix, validate the fact that any node $u$ with ENE $i$ is more likely to have similar ENEs in its neighborhood.

Moreover, we display in Figure~\ref{fig:dist-enes} two plots related to the distribution of ENEs in our dataset.
Notice that the global distribution is heavily skewed: the most present ENE (Person), represents 18\% of the Wikipedia pages in our dataset, whereas at the end of the tail, there are about 10 pages in the dataset.
In other words, we are dealing with extremely imbalanced data, due in part to the nature of the dataset as well as the fine grain of the ENE structure itself. 

However, Figure~\ref{fig:dist-enes-b} shows instead the number of distinct ENEs in the neighbourhood of each page.
In that case, the distribution is much less skewed, as only a limited number of ENEs is present in the neighbourhood of each node.


This analysis serves as a basis to show the relevance of the Wikipedia structure for the classification task.
In order to take into account the structure induced by Wikipedia articles, we present two developments in the following: integrating the graph information with a combination of hand-crafted graph features and natural language features, and through the introduction of a new aggregator for graph neural networks.

Specifically, we experiment with the GraphSAGE model~\cite{hamilton2017inductive}, a recent model of graph neural network that has outperformed state-of-the-art models.
GraphSAGE takes into account the $k$-hop neighbourhood of the graph (typically with $k=2$) to create representations of the nodes that can then be used as features.

    \subsection{Integrating graph structure as features}
 
Let us now define the graph-theoretical features we use.
We use these features for their computability (in polynomial time with respect to the number of nodes), rendering their computation feasible even on large graphs, and because they are classically used as descriptors of real-world graphs. 
Our goal is to describe the structure around each node, in a way that pages with similar roles in the graph will exhibit similar structure.

The degree of a node $u$ is the number of neighbours of $u$: $d(u) = |N(u)|$.

The assortativity is classically defined for an edge $(u,v)$, and quantifies the imbalance between the degrees of nodes $u$ and $v$. Let us suppose, without loss of generality, that $d(v) \geq d(u)$. Then, the mean assortativity for each node is:
$$
    a(u) = \frac{1}{d(u)} \sum_{v\in N(u)} \frac{d(u)}{d(v)}
$$

Communities are groups of nodes exhibiting similar properties. The definition is highly context-dependent, however a common one is that a community is a group of nodes densely connected inside, and with loose connections outside.
Formally, given a graph $G=(V,E)$, a community is group of nodes $X\subseteq V$ such that $\delta(X, X\times X\cap E))$ is high, and $\delta((V, X\times (V \setminus X) \cap E)$ is low, where $\delta$ is the graph density.

For community detection in our setting, we rely on the Louvain algorithm~\cite{blondel2008fast}, which produces a partition of the nodes optimizing modularity, and assign the one-hot encoded version community label to each node.

Betweenness centrality, captures the extent to which a node is {\em central}, or important~\cite{brandes2001faster}.
In this context, the centrality of a node $v$ is defined as the fraction of shortest paths going through $v$:
$$
    {\cal B}(v) = \sum_{u,w\in V} \frac{\varphi(u,w,v)}{\varphi(u,w)}
$$

In other words, a node $v$ has high betweenness centrality in a graph if it connects (through shortest paths) many different pairs of nodes.



\subsection{Taking into account the structure of the ENE hierarchy}

Moreover, even though the classes can be seen as independent of each other in first approximation, they stem from a hierarchical definition.
Indeed, some ENEs are closer to each other than others, for instance ``Country" and ``Region" are similar, whereas ``Country" and ``Food" are not.

This means that even though we are aiming at perfect classification, it is also acceptable to misclassify into a "neighbouring" ENE.

In order to take into account this hierarchy, we introduce a new aggregator for GraphSAGE, \textsc{WMean}:

$$
    h_v^{(k)} = \sigma\left(W\cdot P\cdot \mu\left(h_v^{(k-1)}\cup \{h_u^{(k-1)}: u\in N(v)\}\right)\right) 
$$

where $h_v^{(k)}$ is the vector representation of $v$ at iteration $k$, $W$ is a matrix of weights, $N(v)$ is the neighborhood set of $v$, $\sigma$ is non-linearity, $\mu$ is the statistical mean, and $P$ is a vector of size $N(v)$ weighting differently each neighbour depending on its ENE; we define two scenarios for the vector $P$.

for the sake of simplicity, given a node $v$ and one of its neighbours $u$, $P$ can take the following values: $0.25$ if $ENE(u)\ne ENE(v)$, $0.75$ if $ENE(u)\approx ENE(v)$ and $1$ if $ENE(u) = ENE(v)$.
We say that $ENE(u)\approx ENE(v)$ if the two ENEs have a common parent, {\em i.e.} they belong to the same group.
The goal of this aggregator is to confer more weight to neighbours of $u$ having the same (or a similar) ENE as $u$.
We refer to this scenario as \textsc{WMean-1}.

We also formulate a more flexible function, that, given a hierarchy of classes, ponders the representation inversely proportionally to the distance to the closest common parent between nodes $v$ and $u$, for all $u\in N(v)$.
We refer to this scenario as \textsc{WMean-2}.


\section{Problem setting}

We focus on the problem of accurately classifying unseen Wikipedia pages in different scenarios, in a semi-supervised setting.
In other words, for a given Wikipedia page $u$, we are interest in accurately predicting its label in the ENE hierarchy.

In particular, we are interested in assessing the potential for contribution of the graph structure for this task.
\begin{figure}
    \centering
    \includegraphics[width=0.9\linewidth]{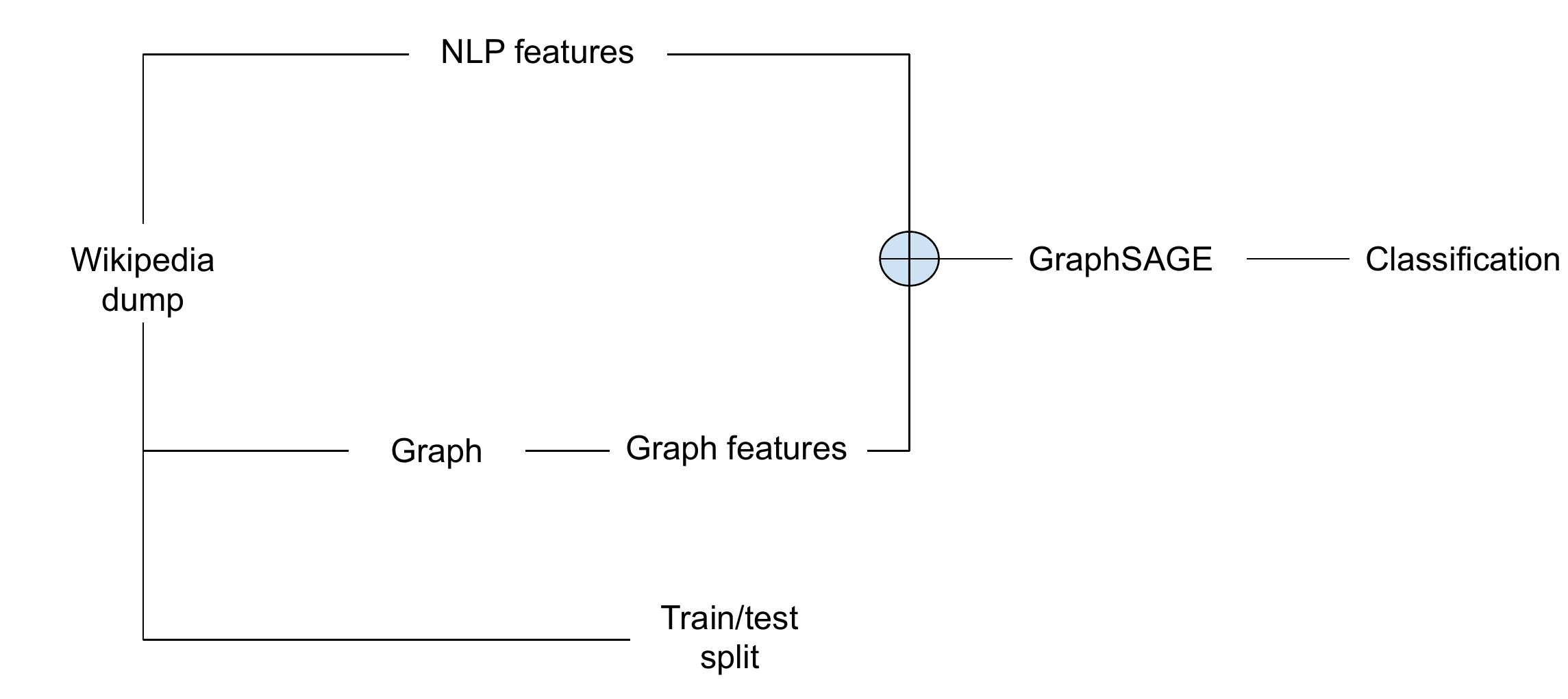}
    \caption{Our experimental setup}
    \label{fig:setting}
\end{figure}

Figure~\ref{fig:setting} summarizes our experimental setting.
We experiment on different features sets, using only NLP features, using graph-based features, and using node embeddings.
Since we are primarily interested in assessing the contribution of graphs, the NLP features are entirely derived from the work described in~\cite{suzuki2016fine}.
We evaluate the prediction accuracy by reporting the micro F1-score.

Before describing the data and our results, let us finally explain how we perform the train/validation/test split.
Indeed, since we are interested in relational data, we believe it is important not to simply select instances at random.
This selection method could lead to disconnected graphs in the test setting, lowering the impact that graph models can have.
To the extreme, the test graph could be completely comprised of isolated nodes.
In order to prevent this, we select training, validation and test in a way that ensures that these graphs are connected (that there is at least a path between all pairs of nodes).

Formally, let us consider a graph $G = (V,E)$, with $V$ a set of nodes and $E\subseteq V\times V$ a set of edges.
We consider $G$ to be simple, i.e. undirected ($(u,v)=(v,u)$), unweighted, and without self-loops ($\forall (u,v)\in E, u\ne v$).

Let us partition the set of nodes $V$ into $3$ sets, $V_{tr}, V_{va}$ and $V_{te}$, respectively the training, validation and testing sets.
These sets are built uniformly at random, such that $\frac{|V_{te}|}{|V|}=0.1$, $\frac{|V_{va}|}{|V|}=0.2$ and $\frac{|V_{tr}|}{|V|}=0.7$.

The training is done on the graph $G_{tr} = (V_{tr}, V_{tr}^2\cap E)$.
In other words, it is the subgraph induced by the training nodes.

The validation is done on the graph $G_{va} = (X, X^2\cap E \setminus V_{va}^2$, where $X=V_{tr}\cup V_{va}$.
This is the graph induced by the training and validation nodes, where all edges between the validation nodes have been removed.

Finally, the testing is done on the graph $G_{te} = (V, E\setminus V_{te}^2)$.
This is the graph where all edges between the test nodes have been removed.

\begin{figure}
    \includegraphics[width=0.3\linewidth]{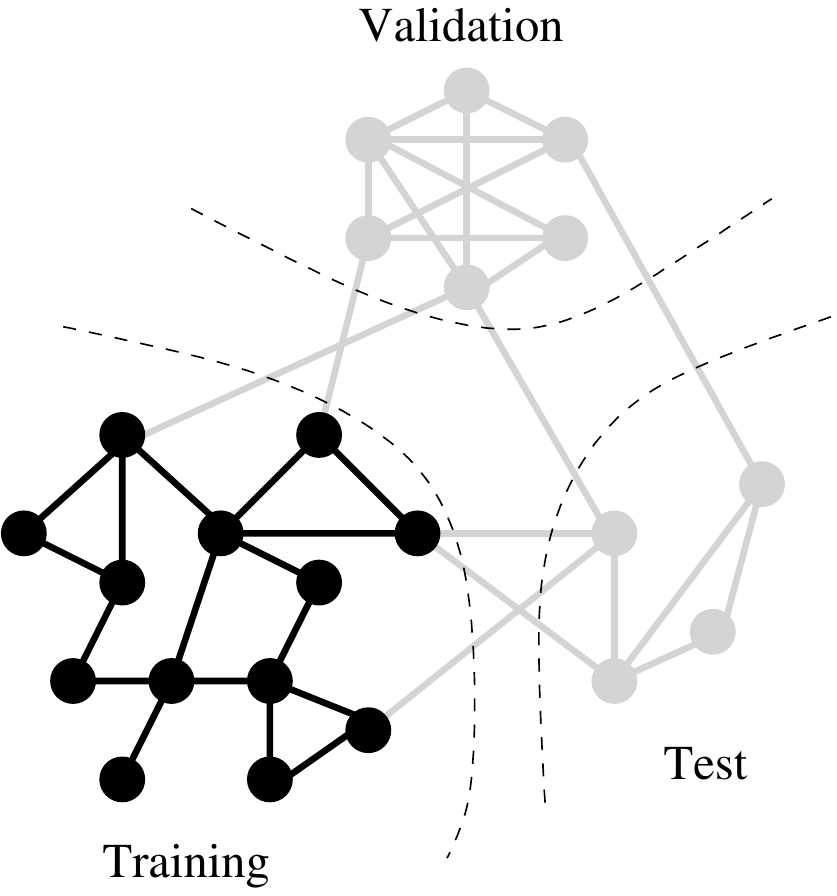}
    \includegraphics[width=0.3\linewidth]{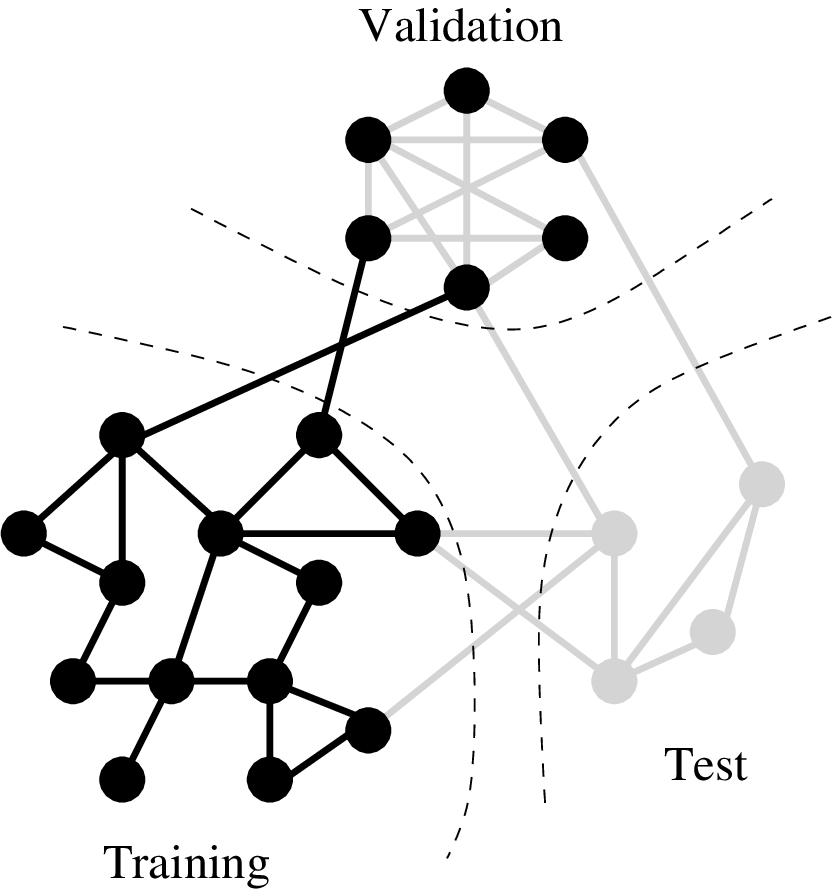}
    \includegraphics[width=0.3\linewidth]{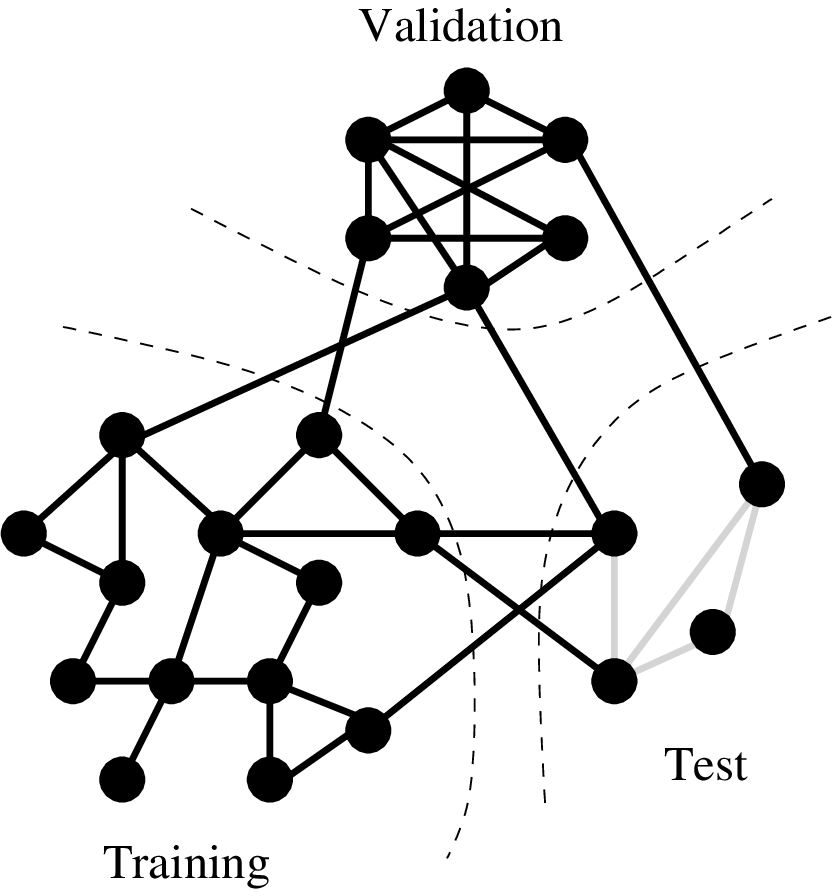}
    \caption{The training, validation and testing graph used in our setting.}
     \label{fig:split-process}
\end{figure}

For an illustration of these three graphs and how they are built, see Figure~\ref{fig:split-process}.

The algorithm only predicts labels for nodes $V_{va}$ in the validation phase, and in $V_{te}$ in the testing phase.

\section{Experiments}

    
    


    \subsection{Results}
    
    Before presenting the results, we tuned the hyperparameters using a random search ($1,000$ trials) on the hyperparameter space.
    The hyperparameter space contains, in addition to the GraphSAGE hyperparameters, the feature set to choose (graph handcrafted features, a subset of the natural language features, node embeddings and their combinations) as well as the aggregator (picked among the four GraphSAGE available ones, or \textsc{Wmean-1} and \textsc{Wmean-2}).
    
    We found that we obtain the best results with two hidden layers, each subsampling respectively $10$, then $25$ times into each node's neighbourhood, and when using a combination of node embeddings (of dimension $128$).

     \begin{table*}
 \centering
\begin{tabular}{|l|c|c|c|c|}
\hline
\multirow{2}{*}{\textbf{Setting}} & \multicolumn{1}{l|}{\multirow{2}{*}{\textbf{\begin{tabular}[c]{@{}l@{}}\# NLP features\\ (max 10,000)\end{tabular}}}} & \multicolumn{1}{l|}{\multirow{2}{*}{\textbf{Aggregator}}} & \multicolumn{2}{c|}{\textbf{Micro F1-score}}                                  \\ \cline{4-5} 
                                  & \multicolumn{1}{l|}{}                                                                                                            & \multicolumn{1}{l|}{}                                     & \multicolumn{1}{l|}{\textbf{Validation}} & \multicolumn{1}{c|}{\textbf{Test}} \\ \hline
\textit{Previous work (reproduced from~\cite{suzuki2016fine})}    & \textit{10,000}                                                                                                                  & \textit{-}                                                & \textit{0.8855}                          & \textit{0.8855}                    \\ \hline\hline
Graph features and node embeddings               & 0                                                                                                                                & MEAN                                                      & 0.52591                                  & 0.53015                            \\ \hline
Natural language features only                 & 5,000                                                                                                                            & MEAN                                                      & 0.86647                                  & 0.84957                            \\ \hline
Node embeddings, graph and natural language features            & 2,000                                                                                                                            & MEAN                                                      & 0.89550                                  & 0.87356                            \\ \hline
Node embeddings and natural language features            & 2,000                                                                                                                            & WMEAN-1                                                     & 0.92526                                  & {\bf 0.92349}                            \\ \hline
Node embeddings and natural language features            & 2,000                                                                                                                            & WMEAN-2                                                    & 0.90441                                  & {\bf 0.91042}                            \\ \hline
\end{tabular}
\caption{Micro F1-score for the ENE classification task. Test results that are better than our baseline are highlighted.}
\label{table:results}
\end{table*}
    
    We present in Table~\ref{table:results} results for the different strategies we elaborated on in Section~\ref{sec:exps} on the dataset presented in Section~\ref{sec:data}.
    Each reported F1-score is the average of a $10$-fold cross-validation.
    

    The first line (in italics) is the result reproduced on this dataset from~\cite{suzuki2016fine}, and is only provided here for reference.
    Notice that it yields slightly better results than presented in~\cite{suzuki2016fine}; this is only due to the use of a more recent version of MeCab, the tokenizer we use to separate words in Japanese, and no other modifications to the methods presented in~\cite{suzuki2016fine} have been made.
    
    From Table~\ref{table:results}, we can see that, first of all, graphs alone are the worst performers (line 2).
    This is expectable, since the task at hand is language-dependent, and the graph structure as we define it has no language related features.
    However, the hybrid model, combining natural language features and graph embeddings (line 4), performs comparably to the baseline, despite using 20\% of the natural language features.
    
    One possible explanation for this stalling may lay into the number of handcrafted graph features, that is very small compared to the number of natural language features, which may lead to their relative lack of impact.

    Finally, our two scenarios \textsc{WMean-1} and \textsc{WMean-2} perform significantly better than our baseline, as reported on lines 5 and 6.

\section{Conclusion}
In this paper, we propose a hybrid classifier to classify articles of the online encyclopedia Wikipedia into a given hierarchy of named entities.
We focus on state-of-the-art models, and assess the interest of integrating  the graph structure in existing classifiers.
We show that indeed, the underlying graph structure is of interest for the classification task.

We then design non-uniform aggregator functions for graph neural networks, and demonstrate that we obtain comparable to slightly better F1-scores using significantly less features (roughly 2,000 instead of 10,000); the computation time is also significantly reduced, from a few hours to less than 10 minutes.

We hope that this work provides a proof of concept of the interest of graph-based features for natural language processing tasks.

This work opens many possibilities for future work, that we briefly detail now.
First of all, and most importantly, we used an older dataset in order to assess the interest of graph models in a setting where comparison to previous work on the same Extended Named Entity ontology was available.
Now that this interest is validated, this opens perspectives to use more recent datasets, that are typically more complete and multilingual.
The Shinra-ML project~\cite{sekine2018shinra} released in October 2019 Wikipedia data in 30 languages, and is as such a dataset of choice for future works on this topic.
In this setting, assessing the relevance of graph-based methods in a multilingual setting (where NLP features are harder to design) is a privileged avenue of work.

Now that comparison with the direct state-of-the-art has been done, another track for future work involves redesigning natural language features. 
Indeed, we suspect that the fact that our performance is hindered by the sheer number of features introduced by~\cite{suzuki2016fine}, as it minimizes the potential for other-sourced features.
In particular, BERT features~\cite{devlin2018bert} is an exciting direction that we intend to follow.

From the graph perspective, we only scratched the surface in terms of what is possible and available.
This includes, of course, new features, and new models.
For example, GraphSAGE, that we base our work upon, heavily relies on neighbourhood sampling to remain computationally efficient even with massive data.
The way this sampling is currently done (uniformly at random) is destructive of outliers and events, which could be important in our task.
Moreover, previous research shows that graphs have a strong potential to make models more explainable~\cite{viard2019augmenting}; in our case, for example, this could mean identifying and visualizing subgraphs associated to classification errors.
We hope that this paper serves as a proof of interest, and can be used as a basis for future works.

\bibliographystyle{named}
\bibliography{main}

\end{document}